\documentclass{article}
\usepackage{spconf,amsmath,graphicx,multicol}
\usepackage{amsfonts}
\usepackage{color}
\usepackage{bbm}
\usepackage{cite}
\usepackage{amssymb}
\usepackage{algorithm}
\usepackage{algorithmic}
\usepackage{tabularx}
\usepackage[utf8]{inputenc}
\usepackage[english]{babel}

\usepackage{amsthm}
\usepackage{subcaption}
\usepackage[font={footnotesize}]{caption}
\usepackage{url}
\usepackage{mathtools}
\usepackage{comment}
\usepackage{float}
\usepackage[keeplastbox]{flushend}

\newtheorem{definition}{Definition}[]

\newtheorem{theorem}{Theorem}[]

\def\y{{\mathbf y}}
\def\s{{\mathbf s}}

\def\W{{\mathbf W}}

\def\C{{\mathbf C}}

\def\P{{\mathbf P}}
\def\R{{\mathbb{R}}}

\def\I{{\mathbf I}}

\def\r{{\mathbf r}}

\def\Y{{\mathbf Y}}
\def\A{{\mathbf A}}

\def\R{{\mathbb{R}}}

\def\I{{\mathbf I}}
\def\a{{\mathbf a}}
\def\t{{\mathbf t}}

\def\u{{\mathbf u}}
\def\c{{\mathbf c}}

\newcommand{\ts}{\textsuperscript}

\let\oldref\ref
\renewcommand{\ref}[1]{(\oldref{#1})}
\newcommand{\RNum}[1]{\uppercase\expandafter{\romannumeral #1\relax}}
\makeatletter
\renewcommand{\fnum@figure}{Fig.~\thefigure}
\makeatother

\title{Accelerated Sparse Subspace Clustering}
\name{Abolfazl Hashemi and Haris Vikalo}
\address{Department of Electrical and Computer Engineering,  University of Texas at Austin, Austin, TX, USA}
\begin{document}
%\ninept
%
\maketitle
\begin{abstract}
State-of-the-art algorithms for sparse subspace clustering perform spectral clustering on a similarity matrix typically obtained by representing each data point as a sparse combination of other points using either basis pursuit (BP) or orthogonal matching pursuit (OMP). BP-based methods are often prohibitive in practice while the performance of OMP-based schemes are unsatisfactory, especially in settings where data points are highly similar. In this paper, we propose a novel algorithm that exploits an accelerated variant of orthogonal least-squares to efficiently find the underlying subspaces. We show that under certain conditions the proposed algorithm returns a subspace-preserving solution. Simulation results illustrate that the proposed method compares favorably with BP-based method in terms of running time while being significantly more accurate than OMP-based schemes. 
\end{abstract}
\vspace{-0.2cm}
%%%%%%%%%%%%%%%%%%%%%%%%%%%%%%%%%%%%%%%%%%%%%%%%%%%%%%%%%%%%%%%%%%
%%%%%%%%%%%%%%%%%%%%%%%%%%%%%%%%%%%%%%%%%%%%%%%%%%%%%%%%%%%%%%%%%%
\section{Introduction}\label{sec:intro}
Massive amounts of data collected by recent information systems give rise to new challenges in the field of signal processing, machine learning, and data analysis.  One such challenge is to develop fast and accurate algorithms so as to find low-dimensional structures in large-scale high-dimensional data sets. The task of extracting such low-dimensional structures is encountered in many practical applications including motion segmentation and face clustering in computer vision \cite{yang2008unsupervised,vidal2008multiframe}, image representation and compression in image clustering \cite{ho2003clustering,hong2006multiscale}, and hybrid system identification in systems theory \cite{vidal2003algebraic}. In these settings, the data can be thought of as being a collection of points lying on a union of low-dimensional subspaces. The goal of subspace clustering is to organize data points into several clusters so that each cluster contains only the points from the same subspace.

Subspace clustering has drawn significant attention over the past decade \cite{vidal2011subspace}. Among various approaches to subspace clustering, methods that rely on spectral clustering \cite{ng2001spectral} to analyze the similarity matrix representing the relations among data points have received much attention due to their simplicity, theoretical rigour, and superior performance. These methods assume that the data is {\em self-expressive} \cite{elhamifar2009sparse}, i.e., each data point can be represented by a linear combination of the other points in the union of subspaces. This motivates the search for a a so-called subspace preserving similarity matrix {\color{black}{which establishes stronger connections among the points originating from a similar subspace}}. To form such a similarity matrix, the sparse subspace clustering (SSC) method in \cite{elhamifar2009sparse,elhamifar2013sparse} employs a sparse reconstruction algorithm referred to as basis pursuit (BP) that aims to minimize an $\ell_1$-norm objective by means of convex optimization approaches such as interior point \cite{kim2007interior} or alternating direction of method of multipliers (ADMM) \cite{boyd2011distributed}. In \cite{dyer2013greedy,you2015sparse}, orthogonal matching pursuit (OMP) is used to greedily build the similarity matrix. Low rank subspace clustering approaches in \cite{lu2012robust,liu2013robust,favaro2011closed,vidal2014low} rely on convex optimization techniques with $\ell_2$-norm and nuclear norm regularizations and find the singular value decomposition (SVD) of the data so as to build the similarity matrix. Finally, \cite{heckel2015robust} presents an algorithm that constructs the similarity matrix through thresholding the correlations among the data points. Performance of self-expressiveness-based subspace clustering schemes was analyzed in various settings. It was shown in \cite{elhamifar2009sparse,elhamifar2013sparse} that when the subspaces are disjoint (independent), the BP-based method is subspace preserving. \cite{soltanolkotabi2012geometric,soltanolkotabi2014robust} take a geometric point of view to further study the performance of BP-based SSC algorithm in the setting of intersecting subspaces and in the presence of outliers. These results are extended to the OMP-based SSC in \cite{dyer2013greedy,you2015sparse}.  
%%%%%%%%%%%%%%%%%%%%%%%%%%%%%%%%%%%%%%%%%%%%%%%%%%%%%%%%%%%%%%
%%%%%%%%%%%%%%%%%%%%%%%%%%%%%%%%%%%%%%%%%%%%%%%%%%%%%%%%%%%%%%

Sparse subspace clustering of large-scale data is computationally challenging. The computational complexity of state-of-the-art BP-based method in \cite{elhamifar2009sparse} and the low rank representation methods \cite{lu2012robust,liu2013robust,favaro2011closed,vidal2014low} is often prohibitive in practical applications. On the other hand, current scalable SSC algorithms, e.g., \cite{dyer2013greedy,you2015sparse}, may produce poor clustering solutions, especially in scenarios where the subspaces {\color{black}{are not well separated}}.
In this paper, we address these challenges by proposing a novel self-expressiveness-based algorithm for subspace clustering that exploits a fast variant of orthogonal least-squares (OLS) to efficiently form a similarity matrix by finding a sparse representation for each data point. We analyze the performance of the proposed scheme and show that in the scenarios where the subspaces are independent, the proposed algorithm always finds a solution that is subspace-preserving. Simulation studies illustrate that our proposed SSC algorithm significantly outperforms the state-of-the-art method \cite{elhamifar2009sparse} in terms of runtime while providing essentially the same or better clustering accuracy. The results further illustrate that, unlike the methods in \cite{elhamifar2009sparse,dyer2013greedy,you2015sparse}, when the subspaces are dependent our
proposed scheme finds a subspace preserving solution.
%\end{enumerate}
%%%%%%%%%%%%%%%%%%%%%%%%%%%%%%%%%%%%%%%%%%%%%%%%%%%%%%%%%%%%%%
%%%%%%%%%%%%%%%%%%%%%%%%%%%%%%%%%%%%%%%%%%%%%%%%%%%%%%%%%%%%%%

The rest of the paper is organized as follows. Section \oldref{sec:pre} formally states the subspace clustering problem and reviews some relevant concepts. In Section \oldref{sec:alg}, we introduce the accelerated sparse subspace clustering algorithm and analyze its performance. Section \oldref{sec:sim} presents the simulation results while the concluding remarks are stated in Section \oldref{sec:concl}. \footnote{The MATLAB implementation of the proposed algorithm is available at \url{https://github.com/realabolfazl/ASSC}.}
\vspace{-0.2cm}
%%%%%%%%%%%%%%%%%%%%%%%%%%%%%%%%%%%%%%%%%%%%%%%%%%%%%%%%%%%%%%%%%%
%%%%%%%%%%%%%%%%%%%%%%%%%%%%%%%%%%%%%%%%%%%%%%%%%%%%%%%%%%%%%%%%%%
\section{Problem Formulation}\label{sec:pre}
%\subsection{Notation}
First, we briefly summarize notation used in the paper and then  formally introduce the SSC problem. 

Bold capital letters denote matrices while 
bold lowercase letters represent vectors. For a matrix $\A$, $\A_{ij}$ denotes the $(i,j)$ entry of
$\A$, and $\a_j$ is the $j\ts{th}$ column of $\A$.
%, and $\A_{-j}$ is the matrix constructed by removing the $j\ts{th}$ column of $\A$. 
Additionally, $\A_S$ is the submatrix of $\A$ that contains the columns of $\A$ indexed by the set $S$. 
${\cal L}_S$ denotes the subspace spanned by the columns of $\A_S$. $\P_S^\bot=\I-\A_S \A_S^\dagger$ is the projection operator 
onto the orthogonal complement of ${\cal L}_S$ where $\A_S^\dagger=\left(\A_S^{\top}\A_S\right)^{-1}\A_S^{\top}$ 
denotes the Moore-Penrose pseudo-inverse of $\A_S$ and $\I$ is the identity matrix. Further, let $[n] = \{1,\dots,n\}$, $\mathbf{1}$ be the vector of all ones, and $\mathcal{U}(0,q)$ denote the uniform distribution on $[0,q]$.

%%%%%%%%%%%%%%%%%%%%%%%Subspace Clustering%%%%%%%%%%%%%%%%%%%%
%\subsection{Sparse Subspace Clustering}\label{sec:ssc}
The SSC problem is detailed next. Let $\{\y\}_{i=1}^N$ be a collection of data points in $\R^D$ and let $\Y = [\y_1,\dots,\y_N] \in \R^{D\times N}$ be the data matrix representing the data points. The data points are drawn from a union of n subspaces $\{S_i\}_{i=1}^n$ with dimensions $\{d_i\}_{i=1}^n$. Without a loss of generality, we assume that the columns of $\Y$, i.e., the data points, are normalized vectors with unit $\ell_2$ norm. The goal of subspace clustering is to partition $\{\y\}_{i=1}^N$ into $n$ groups so that the points that belong to the same subspace are assigned to the same cluster. In the sparse subspace clustering (SSC) framework \cite{elhamifar2009sparse}, one assumes that the data points satisfy the self-expressiveness property formally stated below.
\begin{definition}
\textit{A collection of data points $\{\y\}_{i=1}^N$ satisfies the self-expressiveness property if each data point has a linear representation in terms of the other points in the collection, i.e., there exist a representation matrix $\C$ such that}
\begin{equation}
\Y = \Y \C, \quad \mathrm{diag}(\C) = \mathbf{0}.
\end{equation}
\end{definition}
Notice that since each point in $S_i$ can be
written in terms of at most $d_i$ points in $S_i$, SSC aims to find a sparse subspace preserving 
$\C$ as formalized next.
\begin{definition}
\textit{A representation matrix $\C$ is subspace preserving if for all $j,l \in [N]$ and a 
subspace $S_i$} it holds that
\begin{equation}
\C_{lj} \neq 0 \quad \Longrightarrow \quad \y_j, \y_l \in S_i.
\end{equation}
\end{definition}
The task of finding a subspace preserving $\C$ leads to the optimization problem \cite{elhamifar2009sparse}
\begin{equation}\label{eq:prob}
\begin{aligned}
& \underset{\c_j}{\text{min}}
\quad \|\c_j\|_0
& \text{s.t.}\hspace{0.5cm}  \y_j = \Y\c_j, \quad \C_{jj} = 0,
\end{aligned}
\end{equation}
where $\c_j$ is the $j\ts{th}$ column of $\C$. Given a subspace
preserving solution $\C$, one constructs a similarity matrix $\W = |\C|+|\C|^\top$ for the data points. The graph normalized Laplacian of the similarity matrix $\W$ is then used as an input to a spectral clustering algorithm \cite{ng2001spectral} which in turn produces clustering assignments.
\vspace{-0.2cm}
%%%%%%%%%%%%%%%%%%%%%%%%%%%%%%%%%%%%%%%%%%%%%%%%%%%%%%%%%%%%%%%%%%
%%%%%%%%%%%%%%%%%%%%%%%%%%%%%%%%%%%%%%%%%%%%%%%%%%%%%%%%%%%%%%%%%%
\section{Accelerated OLS for Subspace Clustering}\label{sec:alg}
In this section, we develop a novel self-expressiveness-based algorithm for the subspace clustering problem and analyze its performance.  We propose to find an approximate solution to the problem
\begin{equation}\label{eq:prob-aols}
\begin{aligned}
& \underset{\c_j}{\text{min}}
\quad \|\c_j\|_0
& \text{s.t.}\hspace{0.5cm}  \|\y_j -\Y\c_j\|_2^2 \leq \epsilon, \quad \C_{jj} = 0,
\end{aligned}
\end{equation}
by employing a low-complexity variant of the orthogonal least-squares (OLS) algorithm 
\cite{chen1989orthogonal} so as to find a sparse representation for each data point and 
thus construct $\C$. Note that in \ref{eq:prob-aols}, $\epsilon\geq 0$ is a small predefined parameter that is used as the stopping criterion of the proposed algorithm.

The OLS algorithm, drawn much attention in recent years \cite{hashemi2017sparse,chen1989orthogonal,rebollo2002optimized,hashemi2016sparse,soussen2013joint,herzet2016relaxed}, is a greedy heuristic that iteratively reconstructs sparse signals by identifying one nonzero signal component at a time. The complexity of using classical OLS \cite{chen1989orthogonal} to find a subspace preserving $\C$ -- although lower than that of the BP-based SSC method \cite{elhamifar2009sparse} --  might be prohibitive in applications involving large-scale data. To this end, we propose a fast variant of OLS referred to as accelerated OLS (AOLS) \cite{a2bol2016} that significantly improves both the running time and accuracy of the classical OLS. AOLS replaces the aforementioned single component selection strategy by the procedure where $L\geq 1$ indices are selected at each iteration, leading to significant improvements in both computational cost and accuracy. {\color{black}{To enable significant gains in speed, AOLS efficiently builds a collection of orthogonal vectors $\{\u_{\ell_1},\dots,\u_{\ell_L}\}_{\ell=1}^{T}$ that represent the basis of the subspace that includes the approximation of the sparse signal.}}\footnote{$T < N$ is the maximum number of iterations that depends on the threshold parameter $\epsilon$.}

In order to use AOLS for the SSC problem, consider the task of finding a sparse representation for $\y_j$. Let $A_j\subset[N]\backslash\{j\}$ be the set containing indices of data points with nonzero coefficients in the representation of $\y_j$. That is, for all $l\in A_j$, $\C_{lj}\neq 0$. The proposed algorithm for sparse subspace clustering, referred to as accelerated sparse subspace clustering (ASSC), finds $A_j$ in an iterative fashion (See Algorithm 1). In particular, starting with $A_j=\emptyset$, in the $i\ts{th}$ iteration we identify $L\geq 1$ data points $\{\y_{s_1},\dots,\y_{s_L}\}$ for the representation of $\y_j$. The indices $\{{s_1},\dots,{s_L}\} \subset [N]\backslash(A_j\cup\{j\})$ correspond to the $L$ largest terms $(\y_l^\top \r_{i-1}\slash \y_l^\top \t_l^{(i-1)})^2\|\t_l^{(i-1)}\|_2^2$, where $\r_{i-1} = \P_{A_j}^{\bot}\y_j$ denotes the residual vector in the $i\ts{th}$ iteration with $\r_0 = \y_j$, and 
\begin{equation}
\t_l^{(i)} =\t_l^{(i-1)} -   \sum_{k = 1}^{L}\frac{{\t_l^{(i-1)}}^\top\u_{i_k}}{\|\u_{i_k}\|_2^2}\u_{i_k}
\end{equation}
is the projection of $\y_l$ onto the span of orthogonal vectors 
$\{\u_{\ell_1},\dots,\u_{\ell_L}\}_{\ell=1}^{i}$. Once $\{\y_{s_1},\dots,\y_{s_L}\}$ are selected, we use the assignment
\begin{equation}\label{eq:update}
\u_{i_{k}} = \frac{\y_{s_k}^\top \r_i}{\y_{s_k}^\top \t_{s_k}^{(i)}}\t_{s_k}^{(i)},\quad \r_{i} \leftarrow \r_i - \u_{i_{k}},
\end{equation}
$L$ times to obtain $\r_{i}$ and $\{\u_{\ell_1},\dots,\u_{\ell_L}\}_{\ell=1}^{i}$ that are required for subsequent iterations. This procedure is continued until $\|\r_i\|_2^2< \epsilon$ for some iteration $1\leq i\leq T$, or the algorithm reaches the predefined maximum number of iterations $T$.  Then the vector of coefficients $\c_j$ used for representing $\y_j$ is computed as the least-squares solution $\c_j=\Y_{A_j}^{\dagger}\y_j$. Finally, having found $\c_j$'s, we construct $\W = |\C|+|\C|^\top$ and apply spectral clustering on its normalized Laplacian to obtain the clustering solution.
%=================================== ALGORITHM 1
\renewcommand\algorithmicdo{}	% removes "DO" from for loops
\begin{algorithm}[t]
\caption{Accelerated Sparse Subspace Clustering}
\label{alg:greedy}
\begin{algorithmic}[1]
\STATE \textbf{Input:} $\Y$, $L$, $\epsilon$, $T$\\
\STATE \textbf{Output:} clustering assignment vector $\s$\\
\FOR {$j = 1,\dots,N$}
\STATE Initialize $\r_0 = \y_j$, $i=0$, $A_j = \emptyset$, $\t_l^{0} = \y_l$ for all $l \in [N]\backslash \{j\}$\\
\WHILE  {  $\|\r_i\|_2^2\geq \epsilon$ and $i<T$ }
\STATE Select $\{{s_1},\dots,{s_L}\}$ corresponding to $L$ largest terms
$(\y_l^\top \r_i\slash \y_l^\top \t_l^{(i)})^2\|\t_l^{(i)}\|_2^2$ \\
\STATE $A_j\leftarrow A_j \cup\{{s_1},\dots,{s_L}\}$\\
\STATE $i \leftarrow i+1$\\
\STATE Perform \ref{eq:update} $L$ times to update $\{\u_{\ell_1},\dots,\u_{\ell_L}\}_{\ell=1}^{i}$ and $\r_i$ \\
\STATE $\t_l^{(i)} =\t_l^{(i-1)} -   \sum_{k = 1}^{L}\frac{{\t_l^{(i-1)}}^\top\u_{i_k}}{\|\u_{i_k}\|_2^2}\u_{i_k}$ for all $l \in [N]\backslash \{j\}$\\
\ENDWHILE \\
\STATE $\c_j=\Y_{A_j}^{\dagger}\y_j$\\
\ENDFOR\\
\STATE $\W = |\C|+|\C|^\top$\\
\STATE Apply spectral clustering on the normalized Laplacian of $\W$ to obtain $\s$\\
\end{algorithmic}
\end{algorithm}
%===================================
%%%%%%%%%%%%%%%%%%%%%%%%%%%%%%%%%%%%%%%%%%%%%%%%%%%%%%%%%%%%%%
%%%%%%%%%%%%%%%%%%%%%%%%%%%%%%%%%%%%%%%%%%%%%%%%%%%%%%%%%%%%%%
\vspace{-0.2cm}
\subsection{Performance Guarantee for ASSC}
In this section, we analyze performance of the ASSC algorithm under the scenario that data points are noiseless and drawn from a union of independent subspaces, as defined next.
%===================================
\begin{definition}
\textit{Let $\{S_i\}_{i=1}^{n}$ be a collection of subspaces with dimensions $\{d_i\}_{i=1}^{n}$. Define $\sum_{i=1}^n{S_i} = \{\sum_{i}\y_{i}: \y_{i} \in S_i\}$. Then, $\{S_i\}_{i=1}^{n}$ is called independent if and only if $\mathrm{dim}(\sum_{i=1}^n{S_i}) = \sum_{i=1}^n{d_i}$.}
\end{definition}
% Note that when $n=2$, subspaces are independent if and only if they are disjoint and only intersect at the origin. However, for $n>2$ pairwise disjoint subspaces need not be independent. In addition, any subset of a collection of independent subspaces is also independent. 
%===================================
Theorem \oldref{thm:indep} states our main theoretical results about the performance of the 
proposed ASSC algorithm.
%===================================
\begin{theorem}\label{thm:indep}
\textit{Let $\{\y_i\}_{i=1}^N$ be a collection of noiseless data points drawn  from a union of independent subspaces $\{S_i\}_{i=1}^{n}$. Then, the representation matrix $\C$ returned by the 
ASSC algorithm is subspace preserving.}
\end{theorem}

The proof of Theorem \oldref{thm:indep}, omitted for brevity, relies on the observation that in order to select new representation points, ASSC finds data points that are highly correlated with the current residual vector. Since the subspaces are independent, if ASSC chooses a point that is drawn from a different subspace, its corresponding coefficient will be zero once ASSC meets a terminating criterion (e.g., $\ell_2$-norm of the residual vector becomes less than $\epsilon$ or $T= N-1$). Hence, only the points that are drawn from the same subspace will have nonzero coefficients in the final sparse representation. 
%===================================

\textit{Remark:} It has been shown in \cite{elhamifar2009sparse,dyer2013greedy,you2015sparse} that if subspaces are independent, SSC-BP and SSC-OMP schemes are also subspace preserving. However, as we illustrate in our simulation results, ASSC is very robust with respect to dependencies among the data points across different subspaces while in those settings SSC-BP and SSC-OMP struggle to produce a subspace preserving matrix $\C$. Further theoretical analysis of this setting is left to future work. 
%%%%%%%%%%%%%%%%%%%%%%%%%%%%%%%%%%%%%%%%%%%%%%%%%%%%%%%%%%%%%%
%%%%%%%%%%%%%%%%%%%%%%%%%%%%%%%%%%%%%%%%%%%%%%%%%%%%%%%%%%%%%%
%\subsection{Further Optimization for ASSC}

\vspace{-0.2cm}
%%%%%%%%%%%%%%%%%%%%%%%%%%%%%%%%%%%%%%%%%%%%%%%%%%%%%%%%%%%%%%%%%%
%%%%%%%%%%%%%%%%%%%%%%%%%%%%%%%%%%%%%%%%%%%%%%%%%%%%%%%%%%%%%%%%%%
\section{Simulation Results}\label{sec:sim}
%%%%%%%%%%%%%%%%%%%%%%%%%%%%%%%%%%%%%%%%%%%%%%%%%%%%%%%%%%%%%%%%%%%%%%%%
%%%%%%%%%%%%%%%%%%%%%%%%%%%%%%%%%%%%%%%%%%%%%%%%%%%%%%%%%%%%%%%%%%%%%%%
\begin{figure*}[]
\begin{subfigure}[]{0.24\textwidth}
  \centering
    \includegraphics[width=1\textwidth]{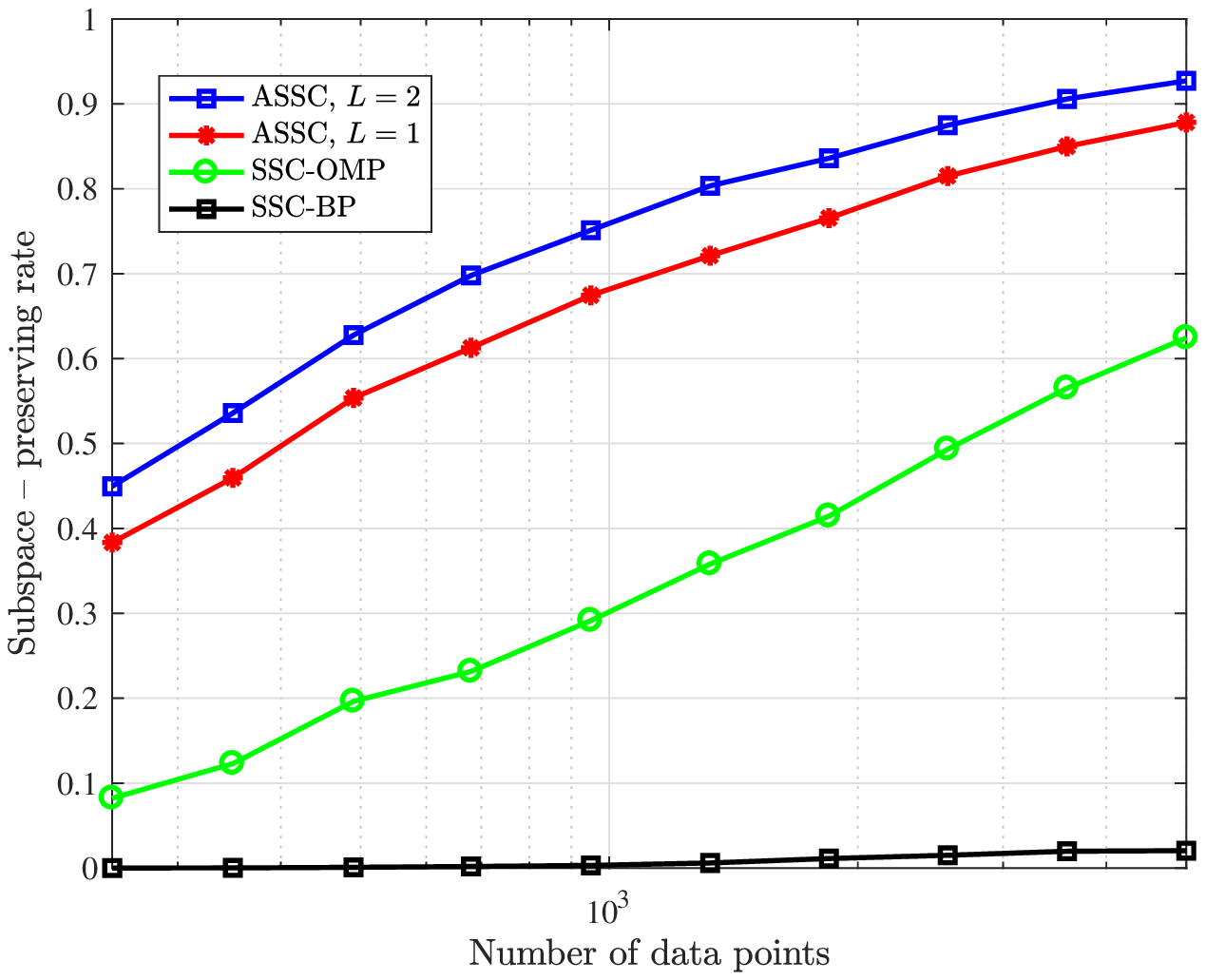}\quad\caption{\footnotesize Subspace preserving rate}
    \end{subfigure}
    \begin{subfigure}[]{.24\textwidth}
  \centering
    \includegraphics[width=1\textwidth]{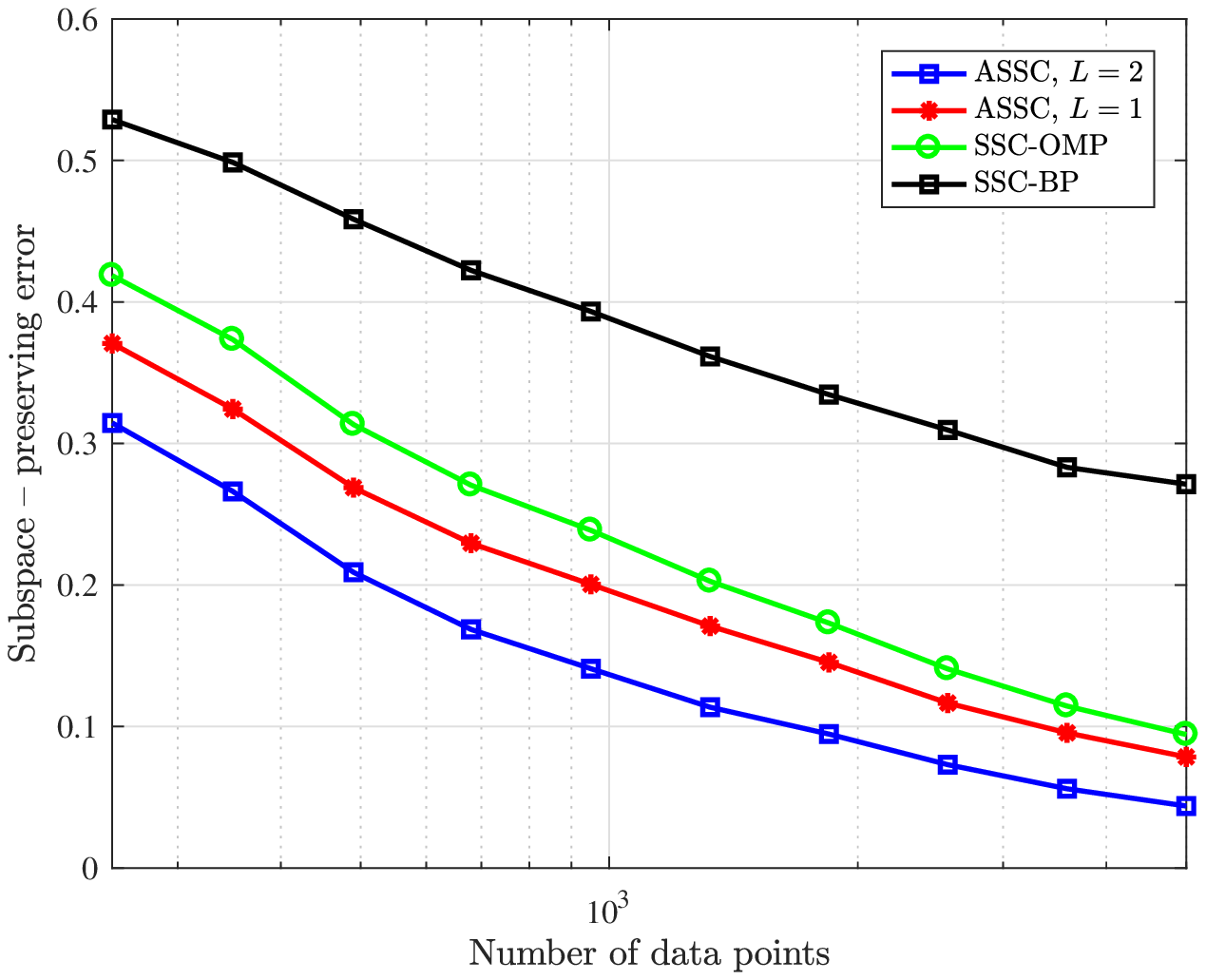}\quad\caption{\footnotesize  Subspace preserving error}
    \end{subfigure}
    \begin{subfigure}[]{.24\textwidth}
  \centering
    \includegraphics[width=1\textwidth]{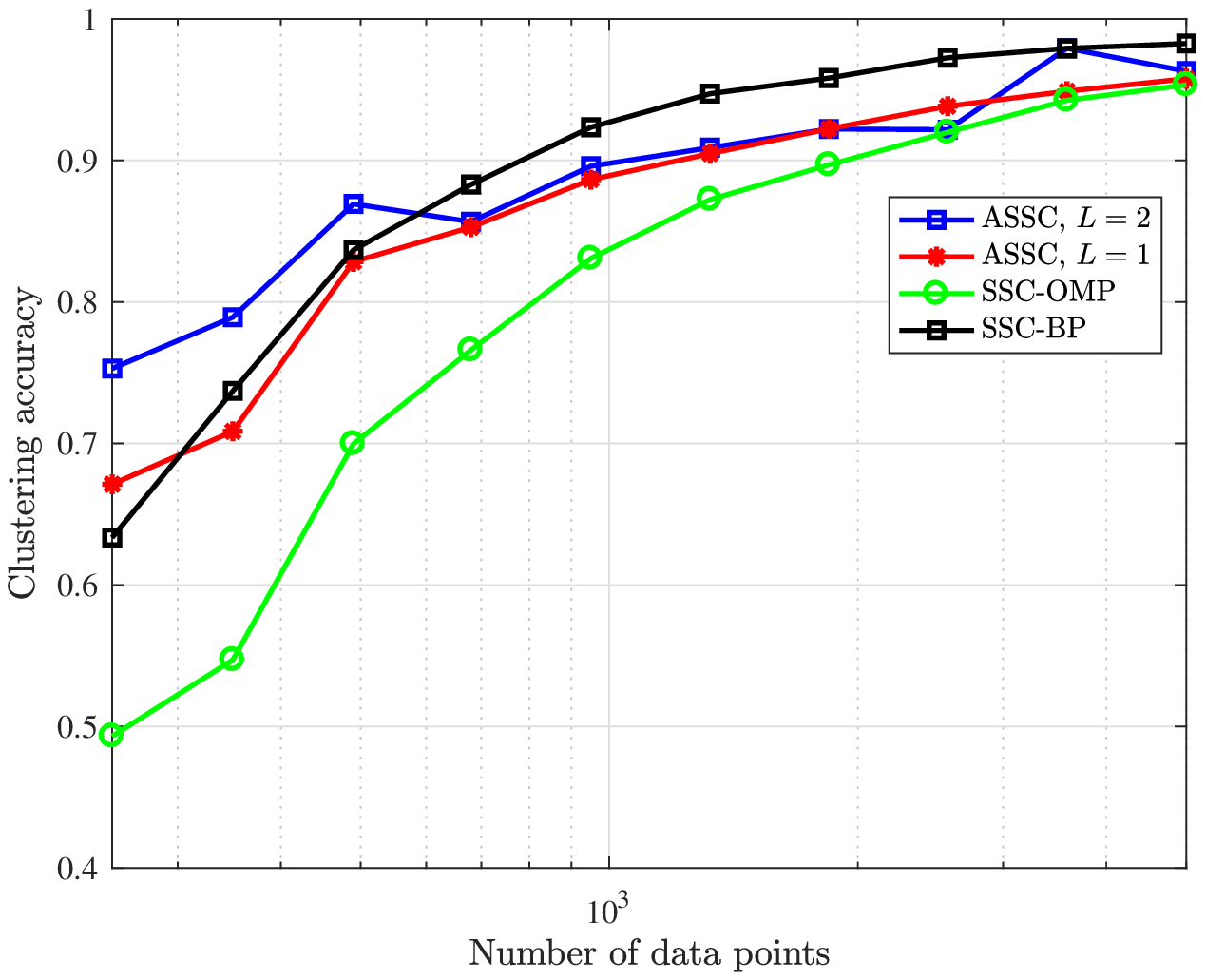}\quad\caption{\footnotesize  Clustering accuracy}
\end{subfigure}
    \begin{subfigure}[]{.24\textwidth}
  \centering
    \includegraphics[width=1\textwidth]{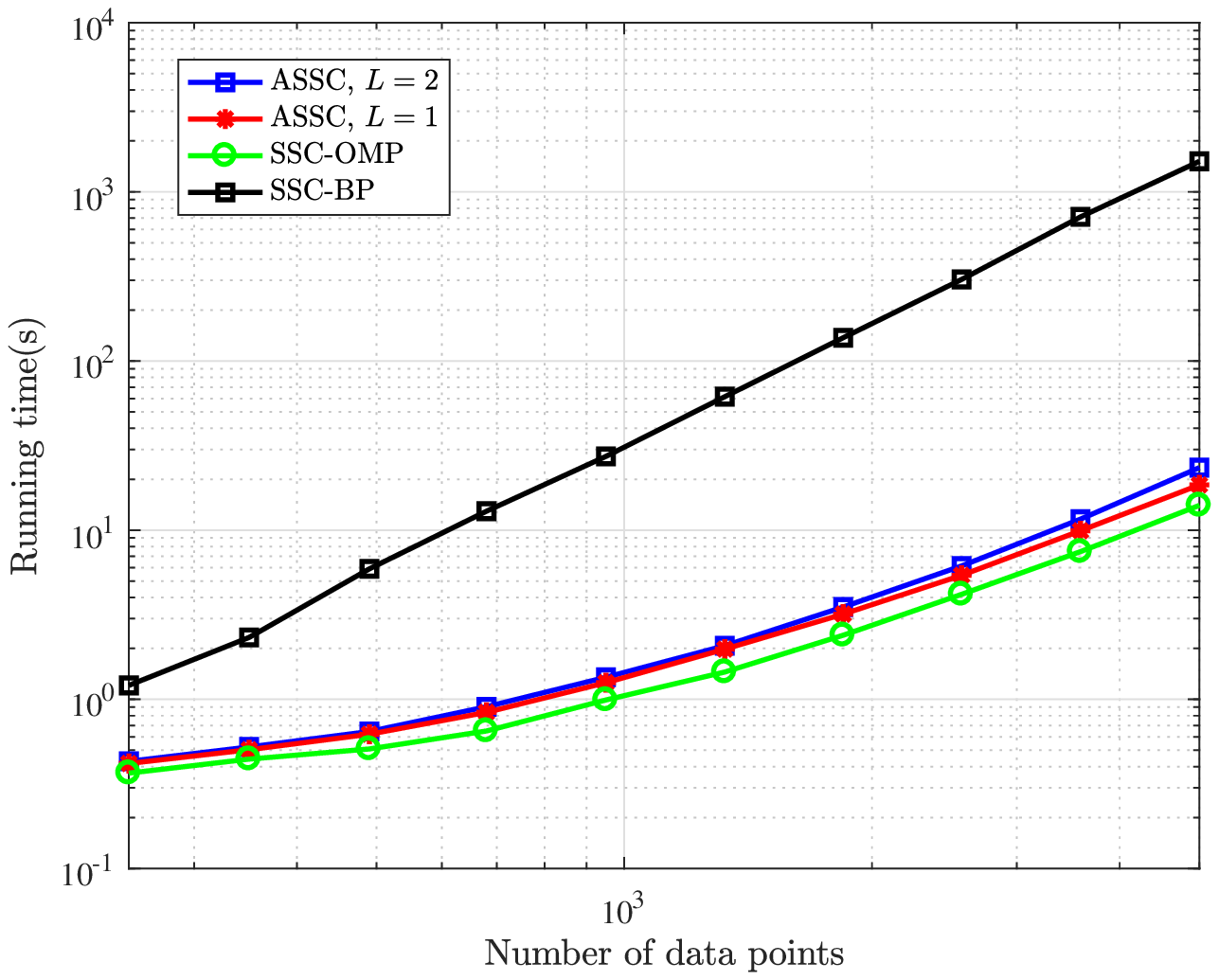}\quad\caption{\footnotesize Running time (sec)}
\end{subfigure}
\caption{\it Performance comparison of ASSC, SSC-OMP \cite{dyer2013greedy,you2015sparse}, and SSC-BP \cite{elhamifar2009sparse,elhamifar2013sparse} {\color{black}{on synthetic data with no perturbation}}. The points are drawn from $5$ subspaces of dimension $6$ in ambient dimension $9$. Each subspace contains the same number of points and the overall number of points is varied from $250$ to $5000$.}
\vspace{-0.1cm}
\end{figure*}
%%%%%%%%%%%%%%%%%%%%%%%%%%%%%%%%%%%%%%%%%%%%%%%%%%%%%%%%%%%%%%%%%%%%%%%%
%%%%%%%%%%%%%%%%%%%%%%%%%%%%%%%%%%%%%%%%%%%%%%%%%%%%%%%%%%%%%%%%%%%%%%%
\begin{figure*}[]
\begin{subfigure}[]{0.24\textwidth}
  \centering
    \includegraphics[width=1\textwidth]{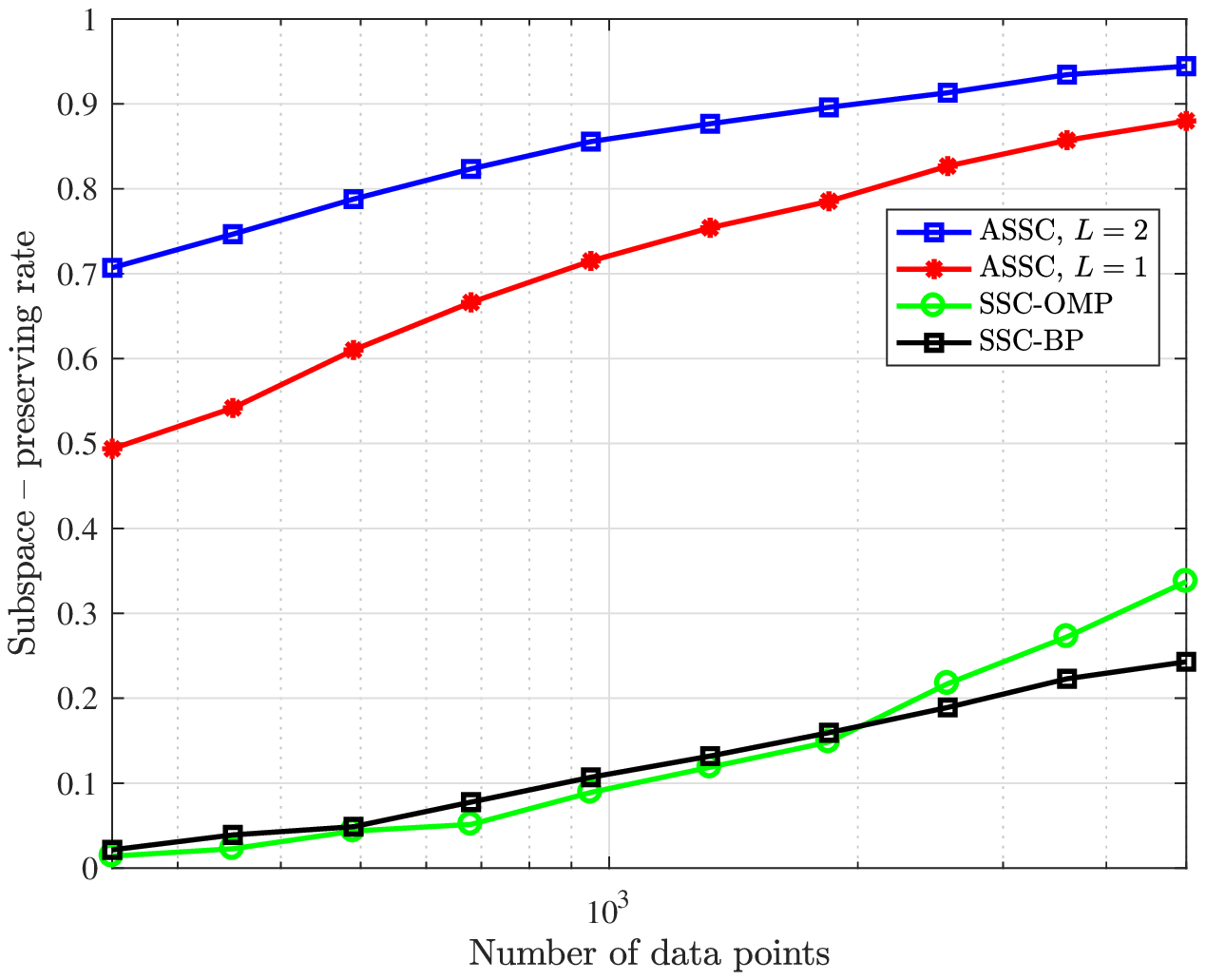}\quad\caption{\footnotesize Subspace preserving rate}
    \end{subfigure}
\begin{subfigure}[]{0.24\textwidth}
  \centering
    \includegraphics[width=1\textwidth]{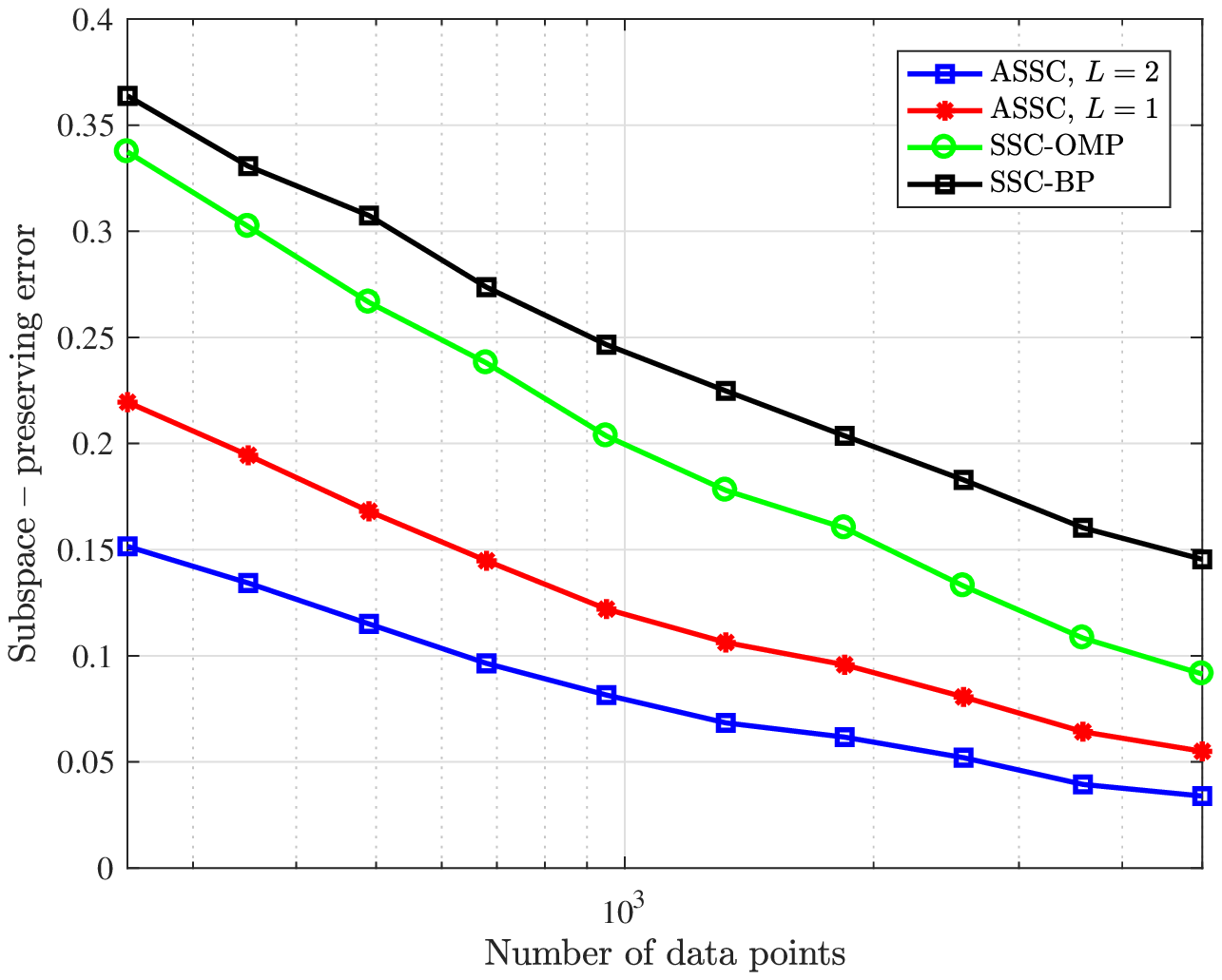}\quad\caption{\footnotesize  Subspace preserving error}
    \end{subfigure}
    \begin{subfigure}[]{.24\textwidth}
  \centering
    \includegraphics[width=1\textwidth]{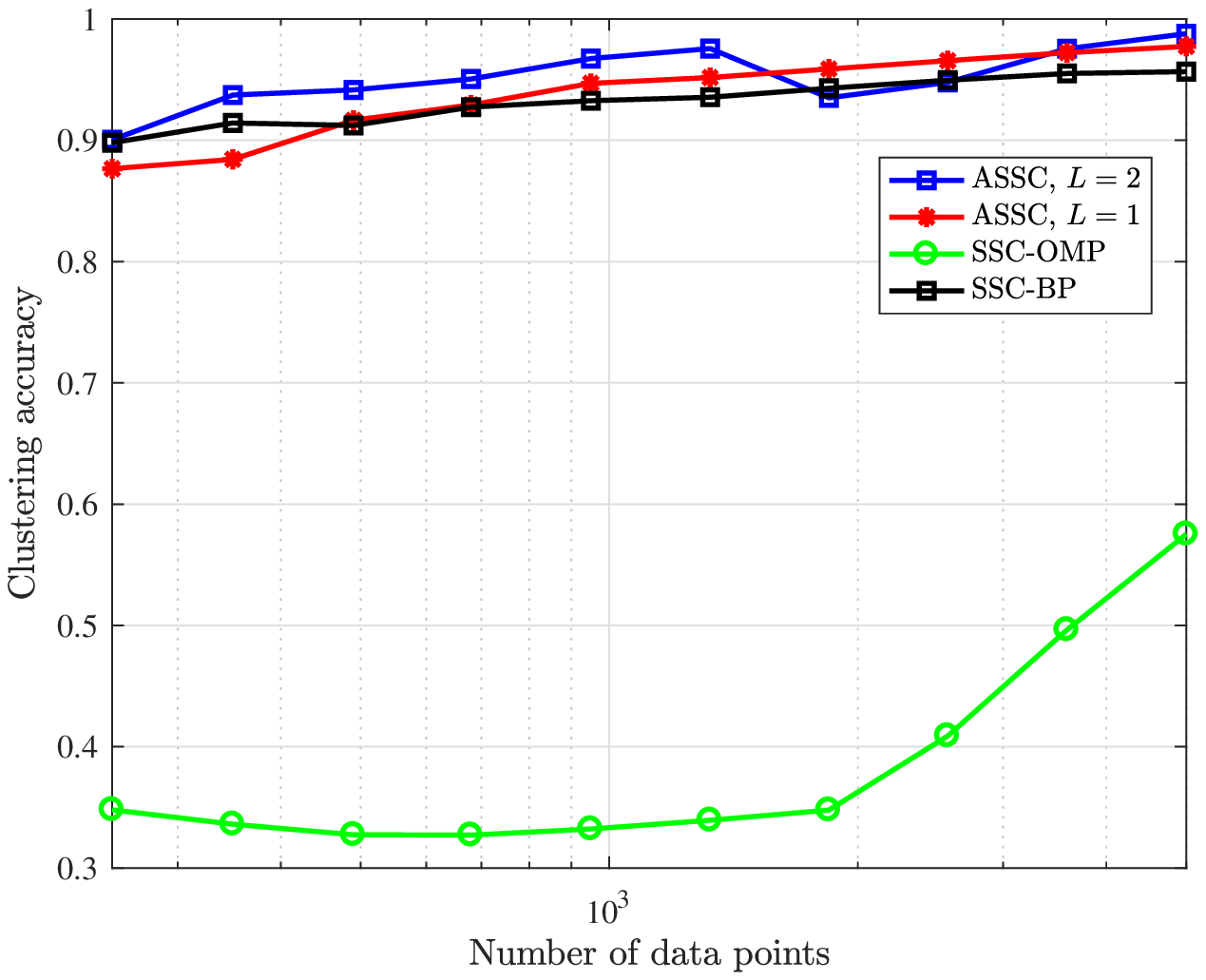}\quad\caption{\footnotesize Clustering accuracy}
    \end{subfigure}
    \begin{subfigure}[]{.24\textwidth}
  \centering
    \includegraphics[width=1\textwidth]{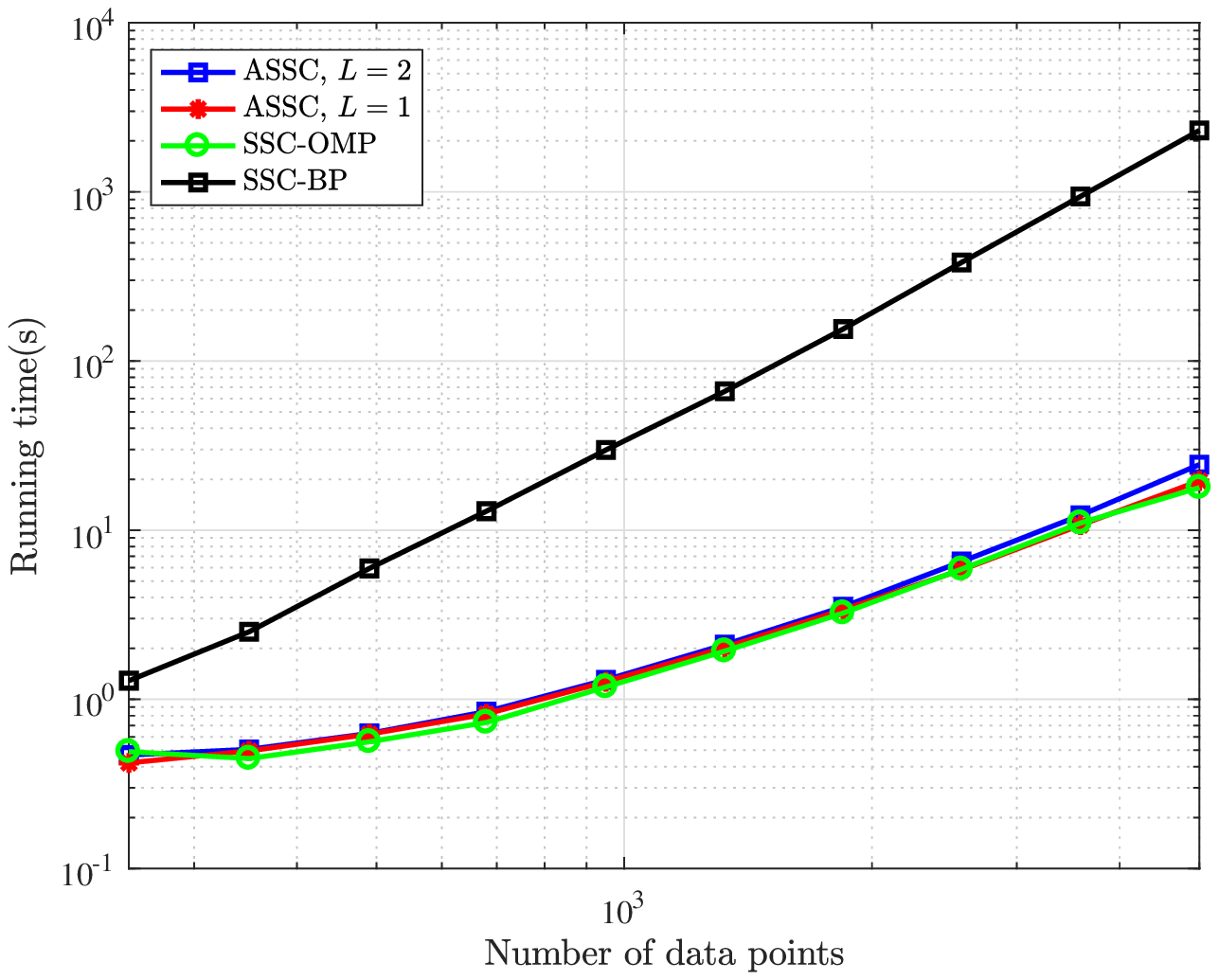}\quad\caption{\footnotesize Running time (sec)}
\end{subfigure}
\caption{\it Performance comparison of ASSC, SSC-OMP \cite{dyer2013greedy,you2015sparse}, and SSC-BP \cite{elhamifar2009sparse,elhamifar2013sparse} {\color{black}{on synthetic data with perturbation terms $Q\sim \mathcal{U}(0,1)$}}. The  points are drawn from $5$ subspaces of dimension $6$ in ambient dimension $9$. Each subspace contains the same number of points and the overall number of points is varied from $250$ to $5000$.}
\vspace{-0.3cm}
\end{figure*}
%%%%%%%%%%%%%%%%%%%%%%%%%%%%%%%%%%%%%%%%%%%%%%%%%%%%%%%%%%%%%%%%%%%%%%%%
%%%%%%%%%%%%%%%%%%%%%%%%%%%%%%%%%%%%%%%%%%%%%%%%%%%%%%%%%%%%%%%%%%%%%%%
To evaluate performance of the ASSC algorithm, we compare it to that of the BP-based \cite{elhamifar2009sparse,elhamifar2013sparse} and OMP-based \cite{dyer2013greedy,you2015sparse} SSC schemes, referred to as SSC-BP and SSC-OMP, respectively. For SSC-BP, two implementations based on ADMM and interior point methods are available by the authors of \cite{elhamifar2009sparse,elhamifar2013sparse}. The interior point implementation of SSC-BP is more accurate than the ADMM implementation while the ADMM implementation tends to produce sup-optimal solution in a few iterations. However, the interior point implementation is very slow even for relatively small problems. Therefore, in our simulation studies we use the ADMM implementation of SSC-BP that is provided by the authors of \cite{elhamifar2009sparse,elhamifar2013sparse}. {\color{black}{Our scheme is tested for $L = 1$ and $L=2$}}. We consider the following two scenarios: (1) A random model where the subspaces are with high probability near-independent; and (2) The setting where we used hybrid dictionaries \cite{soussen2013joint} to generate similar data points across different subspaces which in turn implies the independence assumption no longer holds. In both scenarios, we randomly generate $n = 5$ subspaces, each of dimension $d = 6$, in an ambient space of dimension $D = 9$. Each subspace contains $N_i$ sample points  where we vary $N_i$ from $50$ to $1000$; therefore, the total number of data points, $N = \sum_{i=1}^n N_i$, is varied from $250$ to $5000$. The results are averaged over $20$ independent instances. For scenario (1), we generate data points by uniformly sampling from the unit sphere. For the second scenario, after sampling a data point, we add a perturbation term $Q\mathbf{1}_D$ where $Q\sim \mathcal{U}(0,1)$. 

In addition to comparing the algorithms in terms of their clustering accuracy and running time, we use the following metrics defined in \cite{elhamifar2009sparse,elhamifar2013sparse} that quantify the subspace preserving property of the representation matrix $\C$ returned by each algorithm: \textit{Subspace preserving rate} defined as the fraction of points whose representations are  subspace-preserving, \textit{Subspace preserving error} {\color{black}{defined as the fraction of $\ell_1$ norms of the representation coefficients associated with points from other subspaces, i.e., $\frac{1}{N}\sum_{j}{(\sum_{i\in O}{|\C_{ij}}|\slash \|\c_j\|_1)}$}} where $O$ represents the set of data points from other subspaces.
% \begin{itemize}
% \item \textit{Subspace preserving rate:} The fraction of points whose representations are 
% subspace-preserving.
% \item \textit{Subspace preserving error:} {\color{blue}{The fraction of $\ell_1$ norms of the representation 
% coefficients associated with points from other subspaces, i.e., 
% $\frac{1}{N}\sum_{j}{(1-\sum_{i\in O}{\C_{ij}})\slash \|\c_j\|_1}$}} where $O$ represents the 
% set of data points from other subspaces.
% \end{itemize}

The results for the scenario (1) and (2) are illustrated in Fig. 1 and Fig. 2, respectively. As we see in Fig. 1, ASSC is nearly as fast as SSC-OMP and orders of magnitude faster than SSC-BP while ASSC achieves better subspace preserving rate, subspace preserving error, and clustering accuracy compared to competing schemes. Regarding the second scenario, we observe that the performance of SSC-OMP is severely deteriorated while ASSC still outperforms both SSC-BP and SSC-OMP in terms of accuracy. Further, similar to the first scenario, running time of ASSC is similar to that of SSC-OMP while both methods are much faster that SSC-BP. Overall as Fig. 1 and Fig. 2 illustrate, ASSC algorithm, especially with $L=2$, is superior to other schemes and is essentially as fast as the SSC-OMP method.

\vspace{-0.2cm}
%%%%%%%%%%%%%%%%%%%%%%%%%%%%%%%%%%%%%%%%%%%%%%%%%%%%%%%%%%%%%%%%%%
%%%%%%%%%%%%%%%%%%%%%%%%%%%%%%%%%%%%%%%%%%%%%%%%%%%%%%%%%%%%%%%%%%
\section{Conclusion} \label{sec:concl}
In this paper, we proposed a novel algorithm for clustering high dimensional data lying on a union of subspaces. The proposed algorithm, referred to as accelerated sparse subspace clustering (ASSC), employs a computationally efficient variant of the orthogonal least-squares algorithm to construct a similarity matrix under the assumption that each data point can be written as a sparse linear combination of other data points in the subspaces. ASSC then performs spectral clustering on the similarity matrix to find the clustering solution. We analyzed the performance of the proposed scheme and provided a theorem stating that if the subspaces are independent, the similarity matrix generated by ASSC is subspace-preserving. In simulations, we demonstrated that the proposed algorithm is orders of magnitudes faster than the BP-based SSC scheme \cite{elhamifar2009sparse,elhamifar2013sparse} and essentially delivers the same or better clustering solution. The results also show that ASSC outperforms the state-of-the-art OMP-based method \cite{dyer2013greedy,you2015sparse}, especially in scenarios where the data points across different subspaces are similar. 

As part of the future work, it would be of interest to extend our results and analyze performance of ASSC in the general setting where the subspaces are arbitrary and not necessarily independent. Moreover, it would be beneficial to develop distributed implementations for further acceleration of ASSC.

%%%%%%%%%%%%%%%%%%%%%%%%%%%%%%%%%%%%%%%%%%%%%%%%%%%%%%%%%%%%%%%%%%
%%%%%%%%%%%%%%%%%%%%%%%%%%%%%%%%%%%%%%%%%%%%%%%%%%%%%%%%%%%%%%%%%%
\clearpage
\bibliographystyle{ieeetr}\small
\bibliography{refs}
\end{document}